\documentclass[10pt,twocolumn,letterpaper]{article}

\usepackage{cvpr}  

\definecolor{iccvblue}{rgb}{0.21,0.49,0.74}
\usepackage[pagebackref,breaklinks,colorlinks,allcolors=iccvblue]{hyperref}
\usepackage{graphicx}
\usepackage{amsmath}
\usepackage{amssymb}
\usepackage{colortbl}
\usepackage{graphicx}
\usepackage{xcolor}
\usepackage{colortbl}
\usepackage{multirow}
\usepackage{booktabs} 
\usepackage{amssymb}
\usepackage{pifont}
\usepackage{algorithm}
\usepackage{algpseudocode}

\def\paperID{} 
\def\confName{CVPR}
\def\confYear{2026}

\title{Learning Like Humans: Analogical Concept Learning \\ for Generalized Category Discovery}

\author{Jizhou Han\textsuperscript{\rm 1},
Chenhao Ding\textsuperscript{\rm 2},
Yuhang He\textsuperscript{\rm 1}\thanks{Corresponding author.},
Qiang Wang\textsuperscript{\rm 1},
Shaokun Wang\textsuperscript{\rm 3},\\
SongLin Dong\textsuperscript{\rm 4}, 
Yihong Gong\textsuperscript{\rm 1}
\\[2pt]
\textsuperscript{\rm 1}State Key Laboratory of Human-Machine Hybrid Augmented Intelligence, \\Institute of Artificial Intelligence and Robotics, Xi'an Jiaotong University \\
\textsuperscript{\rm 2}School of Software Engineering, Xi'an Jiaotong University \\
\textsuperscript{\rm 3}Harbin Institute of Technology, Shenzhen \quad
\textsuperscript{\rm 4}Shenzhen University of Advanced Technology 
\\
{\tt\small jizhou-han@stu.xjtu.edu.cn}, 
{\tt\small heyuhang@xjtu.edu.cn}, 
{\tt\small ygong@mail.xjtu.edu.cn}
}

\begin{document}
\maketitle

\begin{abstract}
Generalized Category Discovery (GCD) seeks to uncover novel categories in unlabeled data while preserving recognition of known categories, yet prevailing visual-only pipelines and the loose coupling between supervised learning and discovery often yield brittle boundaries on fine-grained, look-alike categories. We introduce the Analogical Textual Concept Generator (ATCG), a plug-and-play module that analogizes from labeled knowledge to new observations, forming textual concepts for unlabeled samples. Fusing these analogical textual concepts with visual features turns discovery into a visual–textual reasoning process, transferring prior knowledge to novel data and sharpening category separation. ATCG attaches to both parametric and clustering style GCD pipelines and requires no changes to their overall design. Across six benchmarks, ATCG consistently improves overall, known-class, and novel-class performance, with the largest gains on fine-grained data. Our code is available at: \url{https://github.com/zhou-9527/AnaLogical-GCD}.
\end{abstract}

\begin{figure}[t]
  \centering
  \begin{subfigure}{1\linewidth}
    \includegraphics[width=1\linewidth]{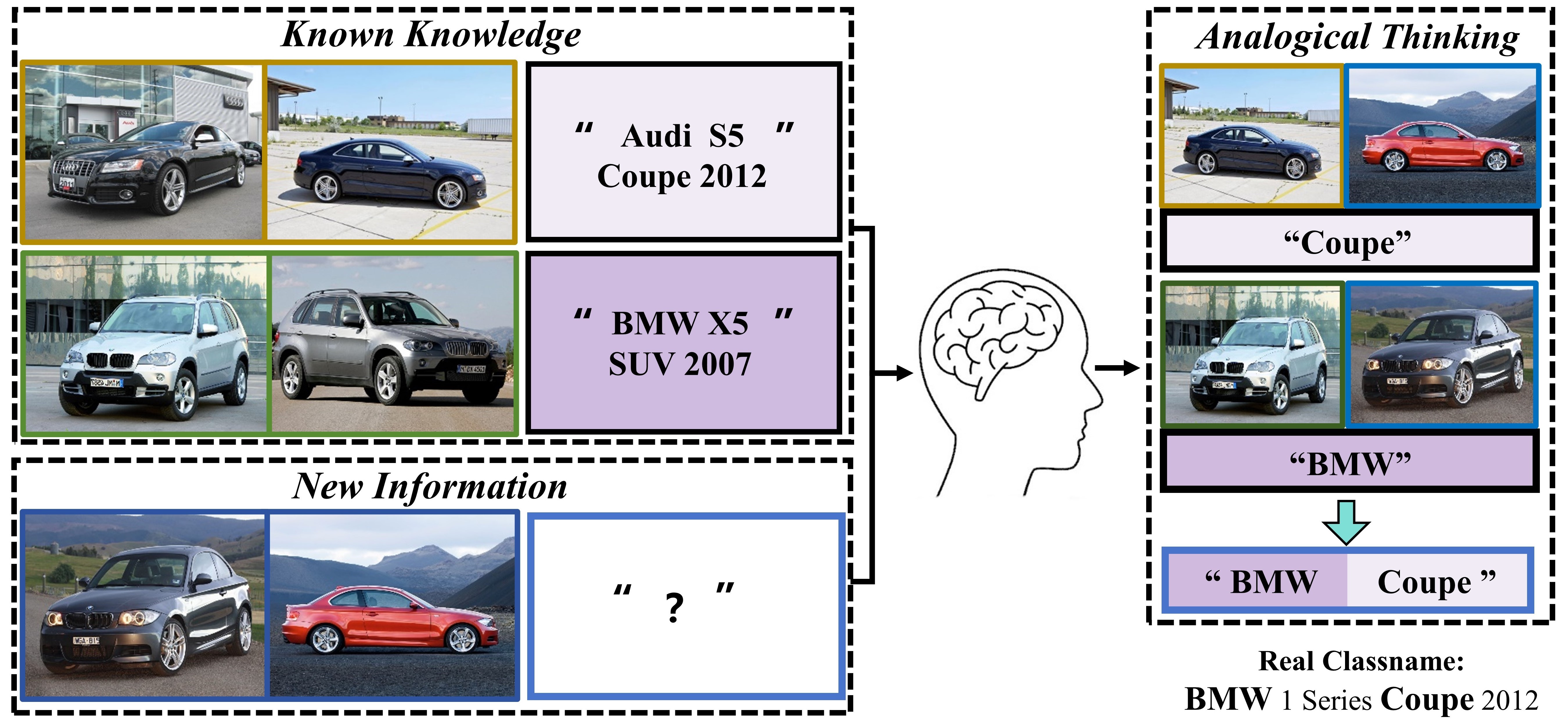}
    \caption{An example of human analogical learning mechanism.}
    \label{fig:short-a}
  \end{subfigure}
  
  \begin{subfigure}{1\linewidth}
    \includegraphics[width=1\linewidth]{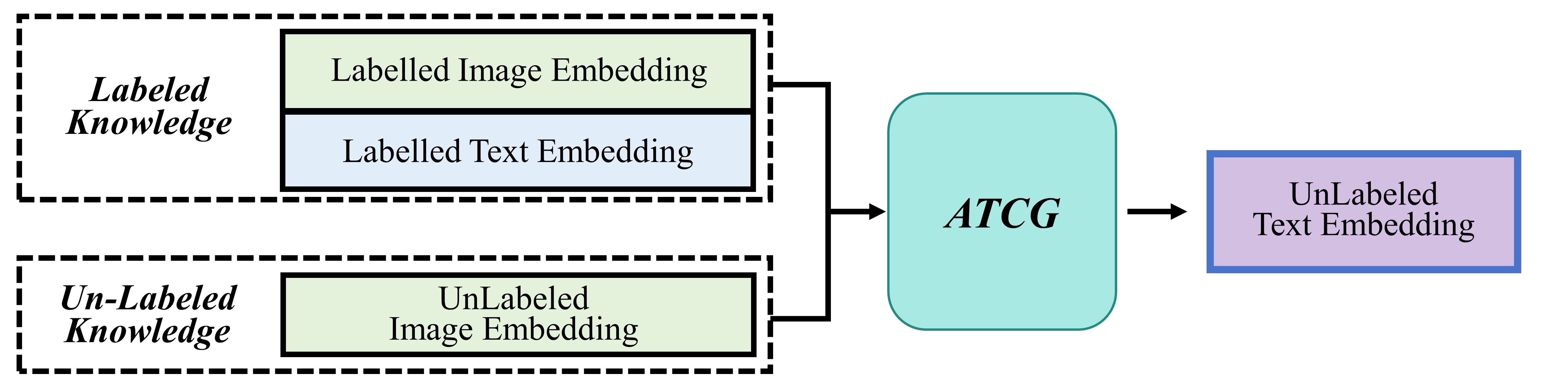}
    \caption{Illustration of the Analogical Textual Concept Generator (ATCG).}
    \label{fig:short-b}
  \end{subfigure}
  \caption{Illustration of an example of human \textbf{Analogical Learning mechanism} and the proposed Analogical Textual Concept Generator (ATCG).}
  \label{fig:short}
\end{figure}

\section{Introduction}
\label{sec:intro}

When applying a well-trained deep neural network to real-world applications, it inevitably encounters data that extends beyond the training data's predefined categories, fundamentally challenging the closed-world assumption underlying traditional learning paradigms. Generalized Category Discovery (GCD)~\cite{GCD2022} addresses this limitation by enabling models to simultaneously recognize previously learned categories while discovering and clustering novel, unseen ones from newly obtained unlabeled data. This capability is essential for deploying intelligent systems in dynamic environments where new concepts or objects naturally emerge.

Recent work on GCD has explored diverse strategies. Early work introduced contrastive learning \cite{GCD2022,DCCL2023,PromptCAL2023} to separate known and unknown categories. Semi-supervised strategies leveraged probabilistic models \cite{GPC2023} for category separation, whereas parametric classification \cite{SimGCD2023} aimed to improve performance but struggled in complex scenarios. In particular, CMS \cite{CMS2024} introduced contrastive mean-shift learning to refine feature representations and also experimented with CLIP by replacing its image encoder. Recently, CLIP-based methods have emerged: CPT \cite{CPT} performs prompt tuning for CLIP with consistency regularization, while GET \cite{GET} adopts a dual-branch design that jointly learns the visual and text branches. 
Despite this progress, most pipelines treat learning from labeled data and discovering novel categories as loosely coupled or even disjoint processes. This decoupling weakens the transfer of prior knowledge to unlabeled data and yields brittle boundaries between visually similar yet semantically distinct categories.

In contrast, humans excel at distinguishing between visually similar categories. This is because, when learning new concepts, humans do not solely rely on visual information. Instead, they utilize contextual knowledge and linguistic cues to form connections between prior knowledge and unknown information, facilitating the creation of new concepts.
Cognitive science research has shown that analogical reasoning plays a central role in human learning and categorization ~\cite{gentner2017analogy,gentner1983structure,holyoak1989analogical}. The hippocampus in the human brain transforms short-term memories into long-term memories stored in the cerebral cortex,  which creates a vast knowledge base \cite{barron2013online}. When encountering a new concept, humans tap into this knowledge base, search for related memories, and use areas such as the prefrontal cortex to draw analogies between new information and existing knowledge \cite{lamprecht2004structural,forbus1995analogical,han2025learn}. This analogy-making process allows humans to abstract common structures and reinterpret information, which promotes learning. As shown in Fig~\ref{fig:short-a}, when learning a new vehicle category, such as ``BMW Coupe," we can efficiently construct the new concept of ``BMW Coupe" by recalling previously acquired relevant concepts, such as ``Audi S5 Coupe" and ``BMW X5 SUV," and drawing analogies between the new image and old concepts.

Inspired by this efficient analogical learning mechanism in human cognition, we propose \textit{Analogical Learning for Generalized Category Discovery (AL-GCD)}, a novel method that mimics the human brain's knowledge retrieval and analogy-making processes. Analogous to how the hippocampus stores and retrieves memories while the prefrontal cortex performs analogical reasoning, AL-GCD constructs a knowledge base from labeled data for knowledge retrieval and leverages it to reason about unlabeled samples through cross-modal analogies. Specifically, AL-GCD comprises four key components: a visual encoder, a text encoder, a fusion module for integrating multimodal representations, and an Analogical Textual Concept Generator (ATCG) that mimics the analogical reasoning process. 
The training procedure consists of two stages. First, in the ATCG training stage, we build a knowledge base that stores concept information from labeled data and train ATCG to generate analogical textual concepts through a pseudo-GCD procedure that simulates concept discovery. Second, in the GCD training stage, as illustrated in Fig.~\ref{fig:short-b}, ATCG processes unlabeled samples by retrieving relevant known concepts and generating analogical textual embeddings for novel categories. These analogical embeddings are then fused with visual features to form integrated representations, enabling effective discovery and recognition of both known and novel categories. 
Extensive experiments on six benchmarks show that AL-GCD outperforms existing methods by \textbf{5.0\%} on average across all datasets and achieves an average overall gain of \textbf{7.1\%} on fine-grained datasets.
In summary, our main contributions include:

\begin{itemize}
\item We introduce an \textbf{Analogical Learning} framework for GCD that bridges existing knowledge and novel-category discovery, enhancing category separation.
\item \textbf{The ATCG module} draws analogies from a labeled visual–textual knowledge base to new observations, producing textual concepts that describe novel samples.
\item Across six benchmarks, ATCG consistently improves overall, known-category, and novel-category performance, with pronounced gains on fine-grained datasets.
\end{itemize}

\begin{figure*}[t]
  \centering
   \includegraphics[width=0.99\linewidth]{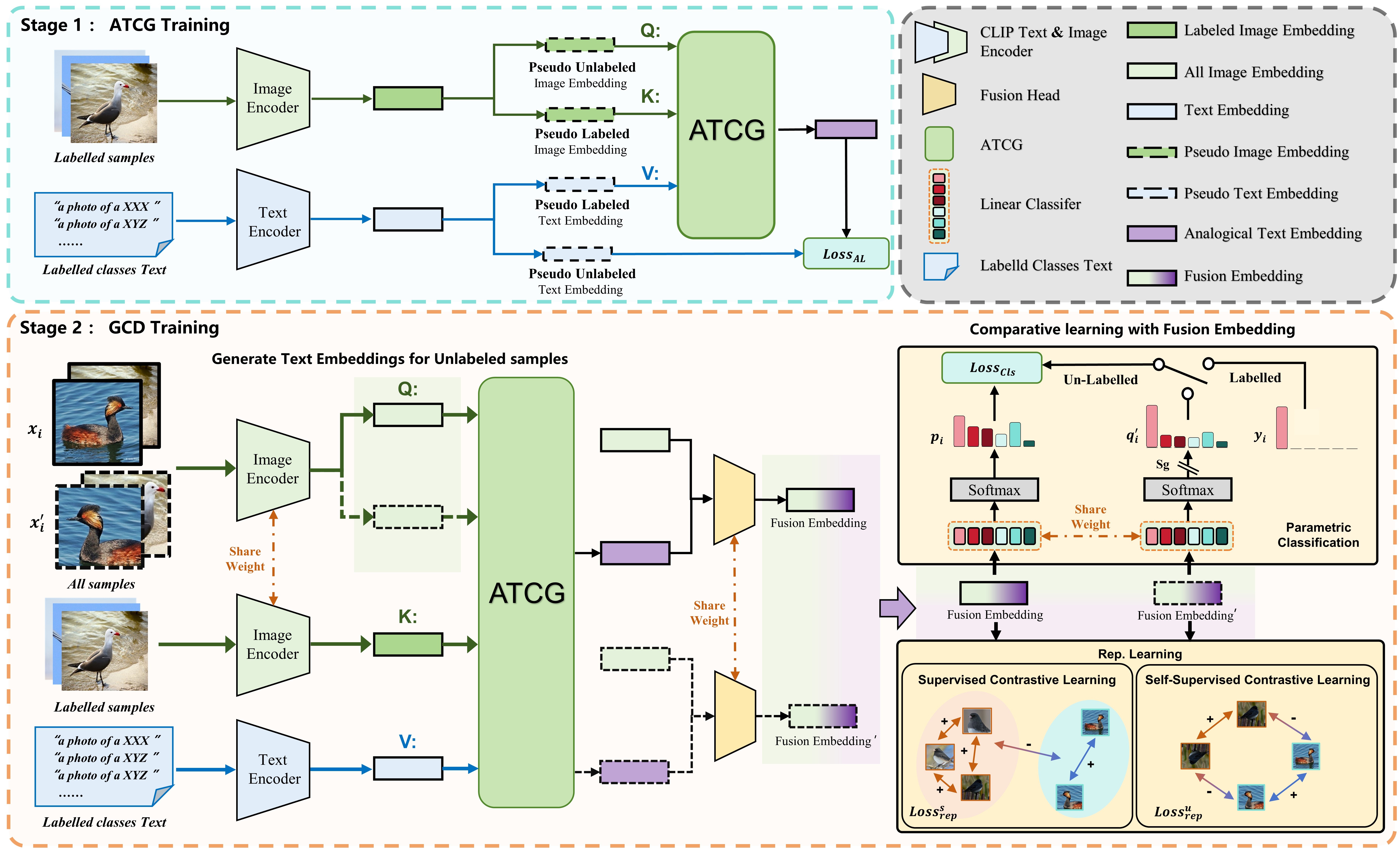}
   \caption{\textbf{Overview of the AL-GCD Framework}. The framework consists of two stages: (1) ATCG training, where the ATCG is trained using labeled images and text embeddings to acquire the ability to generate meaningful analogical text embeddings for unlabeled samples; (2) GCD Training, where ATCG generates text embeddings for unlabeled samples, which are fused with visual embeddings through a fusion head to produce fusion embeddings. These embeddings are optimized via contrastive learning.}
   \label{fig:main}
\end{figure*}

\section{Related Work}
\label{sec:RW}

\subsection{Generalized Category Discovery (GCD)}
GCD seeks to identify novel categories within an unlabeled set while maintaining accuracy on known categories. Early work introduced contrastive learning to separate known and unknown categories \cite{GCD2022}. Subsequent methods refined boundaries via dynamic contrastive learning and prompt-based affinity modeling \cite{DCCL2023, PromptCAL2023}. Semi-supervised approaches leveraged probabilistic modeling with Gaussian mixtures to capture category structure \cite{GPC2023}, whereas parametric classification with category prototypes improved recognition \cite{SimGCD2023}. Further advances emphasized feature enhancement and representation sharpening \cite{PIM2023, AGCD2024}. Some methods~\cite{Xiao2024TRA,NCGCD,han2026goal} impose a preset feature geometry to improve categories separation. CMS \cite{CMS2024} proposed contrastive mean-shift learning to reduce ambiguity and also explored using CLIP by replacing the image encoder. Recently, CLIP-based methods that explicitly leverage language priors have emerged. CPT \cite{CPT} adapts CLIP through consistent prompt tuning to mitigate fine-grained ambiguities. GET \cite{GET} adopts a dual-branch design that jointly learns the visual and text branches, while its final classification still relies on visual embeddings. Despite these advances, many pipelines continue to depend primarily on visual information, limiting their ability to distinguish visually similar yet semantically distinct categories.

\subsection{Vision-Language Models}
In recent years, vision-language models have gained significant attention as powerful tools capable of processing both visual and textual modalities \cite{tan2019lxmert, huang2021seeing,lu2019vilbert}. These models, including CLIP \cite{radford2021learning} and FILIP \cite{yao2021filip}, leverage training on image-text pairs to establish strong connections between vision and language. This allows them to comprehend image semantics while simultaneously understanding their corresponding textual descriptions.
The creation of CLIP marked a milestone, demonstrating how large-scale image-text models can be effectively applied to a wide range of downstream tasks \cite{zhou2022learning,han2025unleashing,zhang2021tip,dong2025beyond,peng2025cia,wang2026structalign}.
Our method integrates textual embeddings with visual features through analogical learning, significantly enhancing model performance and further extending the potential of vision-language models in GCD.

\section{Method}
\label{sec:Method}

\subsection{Problem Setting}
Generalized Category Discovery (GCD) aims to discover novel categories in an unlabeled dataset while maintaining performance on previous categories. Formally, we are given a labeled dataset \( \mathcal{D}^l = \{(x_i^l, y_i^l)\} \subset \mathcal{X} \times \mathcal{Y}^l \) containing samples \( x_i^l \) with labels \( y_i^l \) corresponding to known categories \(\mathcal{Y}^l\), and an unlabeled dataset \( \mathcal{D}^u = \{x_i^u\} \subset \mathcal{X} \) containing samples from both known categories and unknown categories. Here, \( \mathcal{Y}^l \subset \mathcal{Y}^u \), where \( \mathcal{Y}^u \) denotes the set of all categories.
The goal is to train a model to accurately identify unknown categories within \( \mathcal{D}^u \) and integrate them with known categories within a unified classification framework.

\subsection{Overview}
Our framework integrates visual and textual information to enable category discovery through analogical learning. The model comprises a visual encoder $f_v(\cdot)$, a text encoder $f_t(\cdot)$, a Fusion-head projector $g(\cdot)$, and an Analogical Textual Concept Generator (ATCG), denoted by $\varphi_{\text{ATCG}}(\cdot)$. 
As shown in Fig.~\ref{fig:main}, the training pipeline includes two stages: \textbf{ATCG Training} and \textbf{GCD Training}.
In the ATCG Training Phase, we introduce \textit{Knowledge Base Construction} and \textit{Analogical Training}. In the GCD Training Phase, we focus on\textit{ Generating Text Concepts and Fusion Embeddings} and \textit{Contrastive Learning with Fusion Embeddings}. 

During ATCG Training, our approach involves constructing a knowledge base with labeled data, where we extract image and text embeddings to represent each known category. In this phase, we simulate a GCD process called pseudo-GCD, where the ATCG learns to generate textual concepts by analogy, using pseudo-labeled and pseudo-unlabeled data from the knowledge base. In GCD Training, the ATCG generates textual embeddings for unlabeled samples, which are combined with image features to create Fusion Embeddings. These embeddings are optimized using contrastive learning, aiding in category discovery.

\subsection{ATCG Training}
\noindent\textbf{Knowledge Base Construction.} In the ATCG training stage, we first utilize pre-trained visual encoder \( f_v(\cdot) \) and text encoder \( f_t(\cdot) \) to extract features from labeled samples for the construction of the knowledge base. For each sample \( x_i^l \in \mathcal{D}^l \), we obtain an \textit{image embedding} \( \boldsymbol{v}_i^l\) and a \textit{text embedding} \( \boldsymbol{t}_i^l\) as follows:
\begin{equation}
  \boldsymbol{v}_i^l = f_v(x_i^l) \ ,
  \boldsymbol{t}_i^l = f_t(\text{text}(y_i^l)) \
  \label{eq:important}
\end{equation}
where \( \text{text}(y_i^l) \) represents a textual description of the category \( y_i^l \).
The resulting embedding pairs \( \{(\mathbf{v}_i^l, \mathbf{t}_i^l)\} \) are stored in a knowledge base \( \mathcal{K} = \{ (\boldsymbol{v}_i^l, \boldsymbol{t}_i^l) \}_{i \in \mathcal{D}^l} \), which serves as a reference for known categories by encapsulating the relationship between visual and textual representations.

\noindent\textbf{Analogical Training.} Following knowledge base construction, we conduct a pseudo-GCD process to develop ATCG’s ability to generate meaningful text embeddings for unlabeled samples. In each round of training, we randomly divide the labeled dataset \( \mathcal{D}^l \) into pseudo-labeled and pseudo-unlabeled subsets, denoted as \( \mathcal{D}_{\text{P}}^l \) and \( \mathcal{D}_{\text{P}}^u \), respectively. 
Specifically, we split the known categories \( \mathcal{Y}^l \) into pseudo-known categories \(  \mathcal{Y}_{\text{P}}^\text{known} \) and pseudo-unknown categories \(  \mathcal{Y}_{\text{P}}^\text{unknown} \), ensuring \( \mathcal{Y}_{\text{P}}^\text{known} \cap  \mathcal{Y}_{\text{P}}^\text{unknown} = \emptyset \).
Using \( n \) samples from both pseudo-known and pseudo-unknown categories, we create a pseudo-unlabeled set \( \mathcal{D}_{\text{P}}^u \), where each sample is treated as if it were unlabeled, simulating an unknown category. The remaining \( m \) samples from \( \mathcal{Y}_{\text{P}}^\text{known} \) are treated as pseudo-labeled samples, forming the set \( \mathcal{D}_{\text{P}}^l \), with each sample retaining its original label.

For each pseudo-unlabeled sample \( x_j^l \in \mathcal{D}_{\text{P}}^u \), we retrieve its corresponding image and text embeddings \( (\mathbf{v}_j^l, \mathbf{t}_j^l) \) from the knowledge base. Then, the ATCG generates an \textit{Analogical text embedding} \( \tilde{\mathbf{t}}_j \) for each pseudo-unlabeled sample by utilizing its image embedding \( \mathbf{v}_j^l \) along with reference embeddings \( \{ (\mathbf{v}_i, \mathbf{t}_i) \}_{i \in \mathcal{D}_{\text{P}}^l} \) from pseudo-labeled samples:
\begin{equation}
  \tilde{\mathbf{t}}_j = \varphi_{\text{ATCG}}\left(\mathbf{v}_j^l, \{ \mathbf{v}_i \}_{i \in \mathcal{D}_{\text{P}}^l}, \{ \mathbf{t}_i \}_{i \in \mathcal{D}_{\text{P}}^l} \right).
  \label{eq:analogical_text_embedding}
\end{equation}

To ensure that the generated analogical text embeddings \( \tilde{\mathbf{t}}_j \) accurately represent the semantic concepts of the pseudo-unlabeled samples, we define an analogical loss \( \mathcal{L}_{\text{AL}} \):
\begin{equation}
  \mathcal{L}_{\text{AL}} =\frac{1}{n} \sum_{j=1}^{n} \left( 1 - \frac{\tilde{\mathbf{t}}_j \cdot {\mathbf{t}_j^l}^T}{\|\tilde{\mathbf{t}}_j\| \cdot \|\mathbf{t}_j^l\|} \right).
\end{equation}
By iteratively applying this analogical training procedure across various pseudo-labeled and pseudo-unlabeled configurations, the ATCG is optimized to generate conceptually aligned text embeddings effectively. This iterative approach enhances the robustness of ATCG, allowing it to generalize well in generating textual concepts for novel, unseen categories in the actual GCD process.

\subsection{GCD Training}
\noindent\textbf{Generating Text Concepts and Fusion Embedding.} In the GCD Training, we use the trained ATCG to generate meaningful text embeddings for all samples in both \( \mathcal{D}^l \) and \( \mathcal{D}^u \). For each sample \( x_i \in \mathcal{D}^l \cup \mathcal{D}^u \), we obtain an image embedding \(\mathbf{v}_i = f_v(x_i)\).
The ATCG is then employed to generate an analogical text embedding \( \tilde{\mathbf{t}}_i \) for each sample. Using the image embedding \( \mathbf{v}_i \) alongside reference embeddings from the knowledge base \( \mathcal{K} \), the ATCG retrieves relevant visual-textual pairs \( \{ (\mathbf{v}_j^l, \mathbf{t}_j^l) \}_{j \in \mathcal{D}^l} \) and constructs the analogical text embedding as follows:
\begin{equation}
  \tilde{\mathbf{t}}_i = \varphi_{\text{ATCG}}\left(\mathbf{v}_i, \{ \mathbf{v}_j^l \}_{j \in \mathcal{D}^l}, \{ \mathbf{t}_j^l \}_{j \in \mathcal{D}^l} \right).
\end{equation}
Then the generated text embedding $\tilde{t}_i$ is combined with the corresponding image embedding $v_i$ to form a comprehensive fusion embedding $f_i$, enhancing both visual and textual context. The fusion process is defined as follows:
\begin{equation}
    \mathbf{h}_i = \alpha \cdot \mathbf{v}_i + (1 - \alpha) \cdot \tilde{\mathbf{t}}_i
\end{equation}
\begin{equation}
   \mathbf{f}_i = g(\mathbf{h}_i).
\end{equation}
where \( \alpha \) is a balancing coefficient that adjusts the influence of the visual and text embeddings. The intermediate embedding \( \mathbf{h}_i \) is refined through a Fusion-head projection layer \( g(\cdot) \), producing the final Fusion Embedding \( \mathbf{f}_i \).
These embeddings, \( \{ \mathbf{f}_i^l \}_{i \in \mathcal{D}^l} \) and \( \{ \mathbf{f}_i^u \}_{i \in \mathcal{D}^u} \), combine both visual appearance and conceptual insights derived from textual information, aiding in precise category discovery.

\noindent\textbf{Contrastive Learning with Fusion Embeddings.}
In this stage, we refine the fused embeddings \( \mathbf{f}_i^{l} \) and \( \mathbf{f}_i^{u} \) for labeled and unlabeled samples.
ATCG can be plugged into both clustering-based pipelines and approaches with an explicit parametric classifier.
Taking \textit{SimGCD-CLIP \cite{SimGCD2023} + AL-GCD} as an example, the training objective has two components: \textit{Representation Learning} and \textit{Parametric Classification}.

\noindent\textit{Representation Learning.} For all samples, we apply an unsupervised contrastive loss \( \mathcal{L}_{\text{rep}}^{u} \) to maintain consistent representations \cite{GCD2022}. Given a batch \( B \), the unsupervised contrastive loss is defined as:
\begin{equation}
  \mathcal{L}_{\text{rep}}^{u} = -\frac{1}{|B|} \sum_{i \in B} \log \frac{\exp(\mathbf{f}_i \cdot \mathbf{f}_{i}' / \tau)}{\sum_{j \neq i} \exp(\mathbf{f}_i \cdot \mathbf{f}_j / \tau)},
\end{equation}
where \( \mathbf{f}_{i}' \) is an augmented view of \( \mathbf{f}_i \), and \( \tau \) denotes the temperature parameter.
For labeled samples, we use a supervised contrastive loss \( \mathcal{L}_{\text{rep}}^{s} \), grouping samples with identical labels to enhance category separability:
\begin{equation}
  \mathcal{L}_{\text{rep}}^{s} = -\frac{1}{|B_l|} \sum_{i \in B_l} \frac{1}{|P(i)|} \sum_{p \in P(i)} \log \frac{\exp(\mathbf{f}_i^l \cdot \mathbf{f}_p / \tau)}{\sum_{j \neq i} \exp(\mathbf{f}_i^l \cdot \mathbf{f}_j / \tau)},
\end{equation}
where \( P(i) \) represents the set of samples in \( B_l \) with the same label as \( i \).
The combined loss is \( \mathcal{L}_{\text{rep}} = (1 - \lambda) \mathcal{L}_{\text{rep}}^{u} + \lambda \mathcal{L}_{\text{rep}}^{s} \), where \( \lambda \) balances the contributions of unsupervised and supervised components.

\noindent\textit{Parametric Classification.} To further enhance category separation, we incorporate a parametric classification framework as \cite{SimGCD2023}. We initialize category prototypes \( \mathcal{C} = \{ c_1, \dots, c_K \} \) for both known and unknown categories, where \( K = |\mathcal{Y}^u| \). For each sample \( x_i \), the soft label \( p_i(k) \) is computed based on cosine similarity between the Fusion Embedding \( \mathbf{f}_i \) and each prototype \( c_k  \) scaled by \( 1/\tau_s \):
\begin{equation}
  p_i(k) = \frac{\exp \left( \frac{1}{\tau_s} \left( \frac{\mathbf{f}_i}{\|\mathbf{f}_i\|} \cdot \frac{c_k}{\|c_k\|} \right) \right)}{\sum_{k'} \exp \left( \frac{1}{\tau_s} \left( \frac{\mathbf{f}_i}{\|\mathbf{f}_i\|} \cdot \frac{c_{k'}}{\|c_{k'}\|} \right) \right)}.
\end{equation}
We use self-distillation to generate a sharpened pseudo-label $q'_i$ from an alternative view $x'_i$ with temperature $\tau_t$. For labeled samples, the cross-entropy loss is computed between the ground-truth label $y_i$ and the soft prediction $p_i$.
\begin{equation}
  \mathcal{L}_{\text{cls}}^{u} = \frac{1}{|B|} \sum_{i \in B} \mathcal{H}(q'_i, p_i) - \epsilon {H}(\overline{p}),
  \mathcal{L}_{\text{cls}}^{s} = \frac{1}{|B_l|} \sum_{i \in B_l} \mathcal{H}(y_i, p_i)
\end{equation}
where \( \mathcal{H} \) denotes cross-entropy, \( \epsilon \) is a regularization parameter, $\overline{\boldsymbol{p}} = \frac{1}{2|B|}\sum_{i \in B}\left( \boldsymbol{p}_i+\boldsymbol{p}_i^\prime \right)$ denotes the mean prediction of a batch, and the entropy $H(\overline{\boldsymbol{p}}) = -\sum_k\overline{\boldsymbol{p}}^{(k)}\log\overline{\boldsymbol{p}}^{(k)}$. 

The final classification loss is \( \mathcal{L}_{\text{cls}} = (1 - \lambda) \mathcal{L}_{\text{cls}}^{u} + \lambda \mathcal{L}_{\text{cls}}^{s}.\)
The total training objective combines the representation and classification losses \(\mathcal{L} = \mathcal{L}_{\text{rep}} + \mathcal{L}_{\text{cls}}\).

\subsection{Analogical Textual Concept Generator (ATCG)}
\begin{figure}[t]
  \centering
   \includegraphics[width=0.85\linewidth]{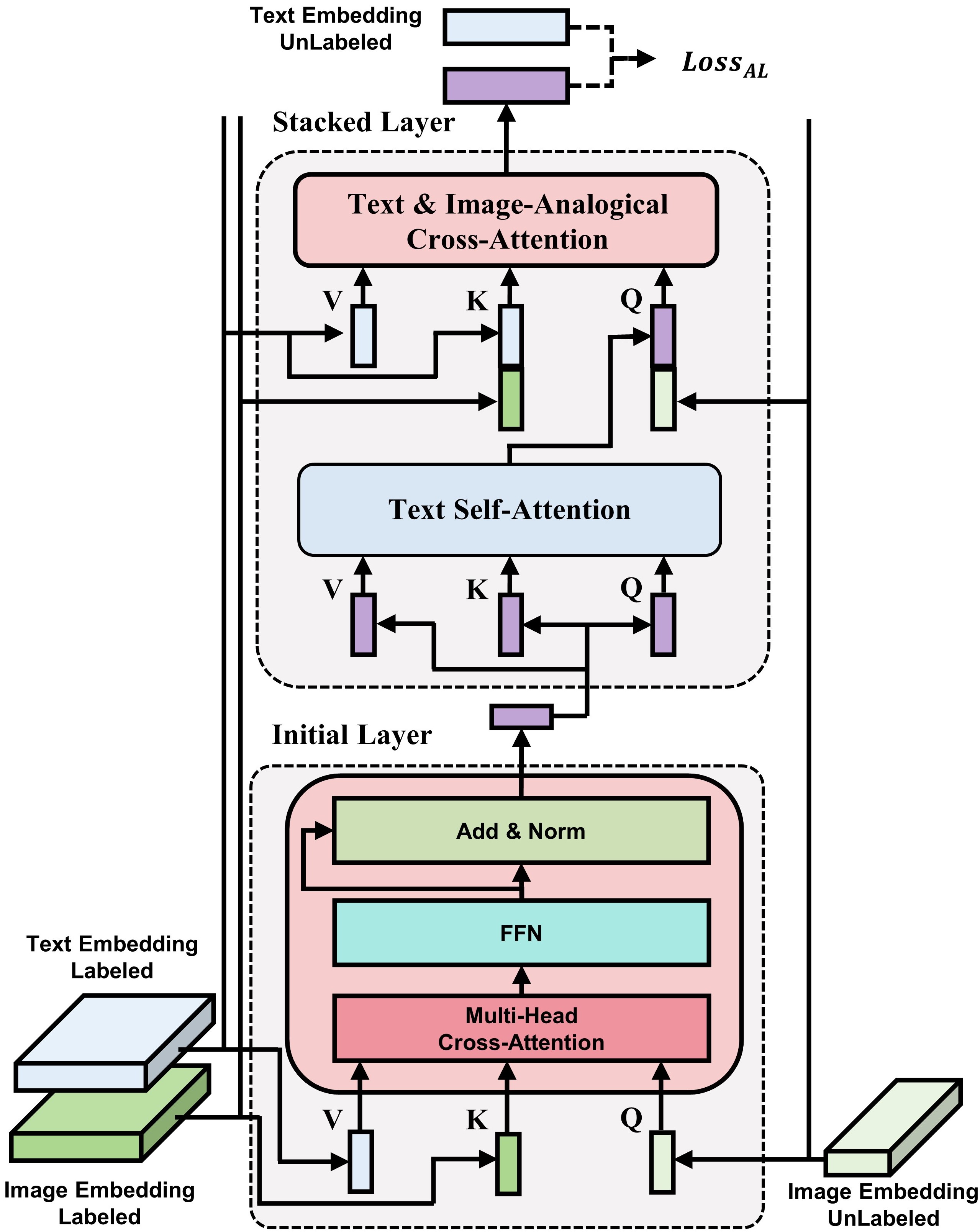}
   \caption{Architecture of the Analogical Textual Concept Generator (ATCG), illustrating its input–output tensor structure.}
   \label{fig:atcg}
\end{figure}

The ATCG is designed to generate analogical text embeddings for all samples, leveraging the relationships between known and unknown categories to produce meaningful textual embeddings. As shown in Fig.~\ref{fig:atcg}, it consists of an \textit{Initial Layer}, which includes only the \textit{Text \& Image-Analogical Attention (TIAA)} module, and multiple \textit{Stacked Layers} for iterative refinement. Each Stacked Layer contains both a \textit{Text Self-Attention (TSA)} module and a \textit{TIAA} module. This layered design allows ATCG to iteratively refine text embeddings, integrating both visual and conceptual insights.

\noindent\textbf{Initial Layer.} In the Initial Layer, the ATCG employs the \textit{Text \& Image-Analogical Attention (TIAA)} module to generate the initial analogical embedding for each unlabeled sample. The components are defined as follows:
\begin{align}
    Q_{\text{A}} = \mathbf{v}_j^u\ , K_{\text{A}} = \{\mathbf{v}_i^l\}_{i \in \mathcal{D}_\text{P}^l}, V_{\text{A}} = \{\mathbf{t}_i^l\}_{i \in \mathcal{D}_\text{P}^l}
\end{align}
where \( Q_{\text{A}} \) is the unlabeled image embedding \( \mathbf{v}_j^u\ \), \( K_{\text{A}} \) is the set of labeled image embeddings, and \( V_{\text{A}} \) is the set of labeled text embeddings.
The initial analogical text embedding \( \tilde{\mathbf{t}}_j^{(0)} \) is obtained by computing the attention between the query and key embeddings and applying it to the value embeddings:
\begin{equation}
    \tilde{\mathbf{t}}_j^{(0)} = \text{softmax} \left( \frac{Q_{\text{A}} \cdot K_{\text{A}}^T}{\sqrt{d}} \right) \cdot V_{\text{A}},
\end{equation}
where \( d \) is the dimensionality of the embeddings.

\begin{table*}[t]
\centering
\small
\setlength{\tabcolsep}{2.4pt}
\caption{Comparison with the state of the art on GCD. The best results are in {bold}, the second best are {underlined}. $^\dagger$ denotes reproduced results. $\triangle$ denotes results from the CMS Appendix \cite{CMS2024}.}
\begin{tabular}{l l|ccc|ccc|ccc|ccc|ccc|ccc}
\toprule
& \multicolumn{1}{l}{\multirow{2}{*}{Method}}
& \multicolumn{3}{c}{CUB}
& \multicolumn{3}{c}{Stanford Cars}
& \multicolumn{3}{c}{FGVC Aircraft}
& \multicolumn{3}{c}{CIFAR100}
& \multicolumn{3}{c}{ImageNet100}
& \multicolumn{3}{c}{Herbarium19}\\
\cmidrule(lr){3-5}\cmidrule(lr){6-8}\cmidrule(lr){9-11}\cmidrule(lr){12-14}\cmidrule(lr){15-17}\cmidrule(lr){18-20}
& & All & Old & New & All & Old & New & All & Old & New & All & Old & New & All & Old & New & All & Old & New \\
\midrule
\multicolumn{20}{l}{\textit{(a) Clustering with the ground-truth number of categories $K$ given}}\\
\midrule
\multirow{6}{*}{\rotatebox{90}{DINO}}
& GCD (CVPR 22)           & 51.3 & 56.6 & 48.7 & 39.0 & 57.6 & 29.9 & 45.0 & 41.1 & 46.9 & 73.0 & 76.2 & 66.5 & 74.1 & 89.8 & 66.3 & 35.4 & 51.0 & 27.0 \\
& GPC (ICCV 23)           & 55.4 & 58.2 & 53.1 & 42.8 & 59.2 & 32.8 & 46.3 & 42.5 & 47.9 & 77.9 & 85.0 & 63.0 & 76.9 & 94.3 & 71.0 &  -   &  -   &  -   \\
& SimGCD (ICCV 23)        & 60.3 & 65.6 & 57.7 & 53.8 & 71.9 & 45.0 & 54.2 & 59.1 & 51.8 & 80.1 & 81.2 & 77.8 & 83.0 & 93.1 & 77.9 & 44.0 & 58.0 & 36.4 \\
& CMS (CVPR 24)           & 68.2 & 76.5 & 64.0 & 56.9 & 76.1 & 47.6 & 56.0 & 63.4 & 52.3 & 82.3 & \textbf{85.7} & 75.5 & 84.7 & 95.6 & 79.2 & 36.4 & 54.9 & 26.4 \\
& SelEx (ECCV 24)         & 73.6 & 75.3 & 72.8 & 58.5 & 75.6 & 50.3 & 57.1 & \underline{64.7} & 53.3 & 82.3 & 85.3 & 76.3 & 83.1 & 93.6 & 77.8 & 39.6 & 54.9 & 31.3 \\

& RLCD (ICML 25)          & 70.0 & 79.1 & 65.4 & 64.9 & 79.3 & 58.0 & \underline{60.6} & 62.2 & \underline{59.8} & 83.4 & 84.2 & 81.9 & 86.9 & 94.2 & 83.2 & 46.4 & 61.2 & 38.4 \\
\cmidrule(lr){2-20}
\multirow{10}{*}{\rotatebox{90}{CLIP}}
& GCD-CLIP $^\triangle$    & 51.1 & 56.2 & 48.6 & 62.5 & 73.9 & 57.0 & 41.2 & 43.0 & 40.2 & \underline{84.2} & 83.1 & \underline{82.3} & 79.3 & 94.6 & 71.1 & 39.7 & 58.0 & 29.9 \\
& CPT (IJCV 25)           & 70.1 & 73.5 & 68.4 & 74.2 & 84.3 & 69.3 & 53.3 & 56.1 & 51.9 & 81.3 & 81.3 & 81.2 & 89.2 & 95.2 & 86.1 & 45.3 & 61.2 & 36.7 \\
& GET (CVPR 25)           & \underline{77.0} & 78.1 & {76.4} & 78.5 & 86.8 & \underline{74.5} & {58.9} & 59.6 & 58.5 & 82.1 & {85.5} & 75.5 & 91.7 & 95.7 & 89.7 & \underline{49.7} & \textbf{64.5} & \underline{41.7} \\
& CMS-CLIP $^\triangle$    & 65.8 & 75.3 & 61.1 & {77.9} & 89.0 & 72.6 & 50.3 & 59.1 & 45.9 & 80.3 & 85.2 & 70.6 & 85.9 & 93.8 & 82.0 & 36.2 & 56.5 & 25.3 \\
& \cellcolor{gray!18} \hspace{2.5em}+ AL-GCD
& \cellcolor{gray!18}73.8 &\cellcolor{gray!18} 79.0 &\cellcolor{gray!18} 71.2 &\cellcolor{gray!18} \textbf{80.0} &\cellcolor{gray!18} \underline{90.3} &\cellcolor{gray!18} \textbf{75.0} &\cellcolor{gray!18} 54.8 &\cellcolor{gray!18} 56.7 &\cellcolor{gray!18} 53.8 &\cellcolor{gray!18} 83.9 &\cellcolor{gray!18} 84.9 &\cellcolor{gray!18} 82.1 &\cellcolor{gray!18} 87.6 &\cellcolor{gray!18} \underline{95.9} &\cellcolor{gray!18} 84.0 &\cellcolor{gray!18} 39.6 &\cellcolor{gray!18} 59.2 &\cellcolor{gray!18} 29.0 \\
& SimGCD-CLIP $^\dagger$   & 69.6 & 75.8 & 66.5 & 69.4 & 82.2 & 62.9 & 53.5 & 57.9 & 51.7 & 81.1 & 84.2 & 75.2 & 89.9 & 94.2 & 87.5 & 47.9 & 62.3 & 39.9 \\
& \cellcolor{gray!18} \hspace{2.5em}+ AL-GCD
        &\cellcolor{gray!18} 74.7 &\cellcolor{gray!18} \underline{79.6} &\cellcolor{gray!18} 72.3 &\cellcolor{gray!18} 78.3 &\cellcolor{gray!18} 86.3 &\cellcolor{gray!18} {74.4} &\cellcolor{gray!18} 58.6 &\cellcolor{gray!18} 61.5 &\cellcolor{gray!18} 57.2 &\cellcolor{gray!18} \textbf{84.7} &\cellcolor{gray!18} 84.5 &\cellcolor{gray!18} \textbf{84.9} &\cellcolor{gray!18} \underline{92.6} &\cellcolor{gray!18} 95.3 &\cellcolor{gray!18} \underline{91.2} &\cellcolor{gray!18} \textbf{50.3} &\cellcolor{gray!18} \underline{63.6} &\cellcolor{gray!18} \textbf{43.1} \\
& SelEx-CLIP $^\dagger$    & 74.2 & 69.5 & \underline{76.5} & 68.8 & 68.4 & 69.0 & 56.7 & 58.2 & 56.0 & 77.2 & 77.0 & 77.6 & 91.2 & 93.6 & 90.0 & 44.4 & 57.1 & 37.6 \\
&\cellcolor{gray!18} \hspace{2.5em}+ AL-GCD
        &\cellcolor{gray!18} \textbf{84.1} &\cellcolor{gray!18} \textbf{79.7} &\cellcolor{gray!18} \textbf{86.3} &\cellcolor{gray!18} \underline{79.0} &\cellcolor{gray!18} \textbf{90.8} &\cellcolor{gray!18} 73.6 &\cellcolor{gray!18} \textbf{66.6} &\cellcolor{gray!18} \textbf{66.1} &\cellcolor{gray!18} \textbf{66.8} &\cellcolor{gray!18} 80.4 &\cellcolor{gray!18} \underline{85.6} &\cellcolor{gray!18} 69.8 &\cellcolor{gray!18} \textbf{92.7} &\cellcolor{gray!18} \textbf{96.2} &\cellcolor{gray!18} \underline{91.1} &\cellcolor{gray!18} 47.7 &\cellcolor{gray!18} 59.0 &\cellcolor{gray!18} 41.6 \\
        
&\cellcolor{gray!32}{\textbf{\textit{Average Improvement}}}
&\cellcolor{gray!32} \textbf{\textit{+7.7}} &\cellcolor{gray!32} \textbf{\textit{+5.9}} &\cellcolor{gray!32} \textbf{\textit{+8.6}}
&\cellcolor{gray!32} \textbf{\textit{+7.1}} &\cellcolor{gray!32} \textbf{\textit{+9.3}} &\cellcolor{gray!32} \textbf{\textit{+6.2}}
&\cellcolor{gray!32} \textbf{\textit{+6.5}} &\cellcolor{gray!32} \textbf{\textit{+3.0}} &\cellcolor{gray!32} \textbf{\textit{+8.1}}
&\cellcolor{gray!32} \textbf{\textit{+3.5}} &\cellcolor{gray!32} \textbf{\textit{+2.9}} &\cellcolor{gray!32} \textbf{\textit{+4.5}}
&\cellcolor{gray!32} \textbf{\textit{+2.0}} &\cellcolor{gray!32} \textbf{\textit{+1.9}} &\cellcolor{gray!32} \textbf{\textit{+2.3}}
&\cellcolor{gray!32} \textbf{\textit{+3.0}} &\cellcolor{gray!32} \textbf{\textit{+2.0}} &\cellcolor{gray!32} \textbf{\textit{+3.6}} \\
\midrule

\multicolumn{20}{l}{\textit{(b) Clustering without the ground-truth number of categories $K$ given}}\\
\midrule
\multirow{3}{*}{\rotatebox{90}{DINO}}
& GCD (CVPR 22)           & 51.1 & 56.4 & 48.4 & 39.1 & 58.6 & 29.7 &  -   &  -   &  -   & 70.8 & 77.6 & 57.0 & 77.9 & 91.1 & 71.3 & 37.2 & 51.7 & \underline{29.4} \\
& GPC (ICCV 23)           & 52.0 & 55.5 & 47.5 & 38.2 & 58.9 & 27.4 & 43.3 & 40.7 & \underline{44.8} & 75.4 & \underline{84.6} & 60.1 & 75.3 & 93.4 & 66.7 & 36.5 & 51.7 & 27.9 \\
& CMS (CVPR 24)           & 64.4 & 68.2 & \underline{62.2} & 51.7 & 68.9 & 43.4 & \underline{55.2} & \textbf{60.6} & \underline{52.4} & \underline{79.6} & 83.2 & \underline{72.3} & 81.3 & \underline{95.6} & 74.2 & 37.4 & 56.5 & 27.1 \\
\cmidrule(lr){2-20}
\multirow{3}{*}{\rotatebox{90}{CLIP}}
& CMS-CLIP (CVPR 24)    & \underline{65.6} & \underline{74.0} & 61.3 & \underline{77.2} & \underline{87.3} & \underline{72.3} & 50.6 & 52.8 & 49.5 & 78.0 & 81.2 & 71.5 & \underline{84.8} & 93.8 & \underline{80.2} & \underline{38.8} & \underline{57.7} & 28.6 \\
&\cellcolor{gray!18} \hspace{2.5em}+ AL-GCD   &\cellcolor{gray!18} \textbf{73.7} &\cellcolor{gray!18} \textbf{79.1} &\cellcolor{gray!18} \textbf{71.0} &\cellcolor{gray!18} \textbf{79.9} &\cellcolor{gray!18} \textbf{90.5} &\cellcolor{gray!18} \textbf{74.7} &\cellcolor{gray!18} \textbf{55.3} &\cellcolor{gray!18} \underline{58.9} &\cellcolor{gray!18} \textbf{53.5} &\cellcolor{gray!18} \textbf{83.3} &\cellcolor{gray!18} \textbf{84.9} &\cellcolor{gray!18} \textbf{80.2} &\cellcolor{gray!18} \textbf{85.7} &\cellcolor{gray!18} \textbf{96.5} &\cellcolor{gray!18} {\textbf{80.3}} &\cellcolor{gray!18} \textbf{41.2} &\cellcolor{gray!18} \textbf{59.0} &\cellcolor{gray!18} \textbf{31.6} \\
&\cellcolor{gray!32}  \textbf{\textit{Improv.\ over CMS-CLIP}}
&\cellcolor{gray!32} \textbf{\textit{+8.1}} &\cellcolor{gray!32} \textbf{\textit{+5.1}} &\cellcolor{gray!32} \textbf{\textit{+9.7}}
&\cellcolor{gray!32} \textbf{\textit{+2.7}} &\cellcolor{gray!32} \textbf{\textit{+3.2}} &\cellcolor{gray!32} \textbf{\textit{+2.4}}
&\cellcolor{gray!32} \textbf{\textit{+4.7}} &\cellcolor{gray!32} \textbf{\textit{+6.1}} &\cellcolor{gray!32} \textbf{\textit{+4.0}}
&\cellcolor{gray!32} \textbf{\textit{+5.3}} &\cellcolor{gray!32} \textbf{\textit{+3.7}} &\cellcolor{gray!32} \textbf{\textit{+8.7}}
&\cellcolor{gray!32} \textbf{\textit{+0.9}} &\cellcolor{gray!32} \textbf{\textit{+2.7}} &\cellcolor{gray!32} \textbf{\textit{+0.1}}
&\cellcolor{gray!32} \textbf{\textit{+2.4}} &\cellcolor{gray!32} \textbf{\textit{+1.3}} &\cellcolor{gray!32} \textbf{\textit{+3.0}} \\
\bottomrule
\end{tabular}
\label{tab:main}
\end{table*}

\noindent\textbf{Stacked Layers.} 
Following the Initial Layer, the ATCG utilizes several Stacked Layers to iteratively refine the analogical text embedding. Each Stacked Layer consists of two key components: the \textit{Text Self-Attention (TSA)} module and an additional  \textit{TIAA} module.

The \textit{TSA} module enhances the analogical embedding by focusing on internal textual coherence. Given the text embedding \( \tilde{\mathbf{t}}_j^{(n-1)} \) from the previous layer, the \textit{TSA} computes:
\begin{align}
    Q_{\text{TSA}} = K_{\text{TSA}} = V_{\text{TSA}} = \tilde{\mathbf{t}}_j^{(n-1)},
\end{align}
\begin{equation}
    \tilde{\mathbf{t}}_j^{\text{TSA}} = \text{softmax} \left( \frac{Q_{\text{TSA}} \cdot K_{\text{TSA}}^T}{\sqrt{d}} \right) \cdot V_{\text{TSA}}.
\end{equation}
The \textit{TSA} encourages the embedding to self-align, refining it to be more contextually coherent within the textual domain.

Following \textit{TSA}, the \textit{TIAA} module in each Stacked Layer incorporates the updated text embedding with the image embedding of the unlabeled sample, enabling a comprehensive analogy. The components for TIAA in the Stacked Layer are defined as:
\begin{align}
    Q_{\text{TIAA}} &= \text{Concat}[\tilde{\mathbf{t}}_j^{\text{TSA}}, \mathbf{v}_j^u], \\
    K_{\text{TIAA}} &= \text{Concat}[\{\mathbf{t}_i^l\}_{i \in \mathcal{D}^l}, \{\mathbf{v}_i^l\}_{i \in \mathcal{D}^l}], \\
    V_{\text{TIAA}} &= \{\mathbf{t}_i^l\}_{i \in \mathcal{D}^l},
\end{align}
The refined analogical text embedding \( \tilde{\mathbf{t}}_j^{n} \) for the next iteration is given by:
\begin{equation}
    \tilde{\mathbf{t}}_j^{n} = \text{softmax} \left( \frac{Q_{\text{TIAA}} \cdot K_{\text{TIAA}}^T}{\sqrt{2d}} \right) \cdot V_{\text{TIAA}}.
\end{equation}

Through \( n \) layers of iteration, ATCG calculates the analogy (similarity) between unlabeled and labeled samples, re-organizing known textual concepts to generate aligned text embeddings \( \tilde{\mathbf{t}}_j^{\text{final}} \) for unlabeled samples. This iterative process effectively guides the discovery of new categories within the GCD framework.

\section{Experiment}
\label{sec:Experiment}

\subsection{Experimental Setup}
\noindent \textbf{Datasets.} Our evaluation spans six image classification benchmarks, including both generic and fine-grained datasets. We use CIFAR-100~\cite{krizhevsky2009learning} and ImageNet-100~\cite{deng2009imagenet} for generic datasets. For fine-grained datasets, we evaluate on CUB-200~\cite{wah2011caltech}, Stanford Cars~\cite{krause20133d}, FGVC Aircraft~\cite{maji2013fine}, and Herbarium19~\cite{tan2019herbarium}. 
To separate categories into known and unknown categories, we follow the SSB split protocol~\cite{li2020ssb} for the fine-grained datasets. For CIFAR-100 and ImageNet-100, we perform a random category split using a fixed seed, consistent with previous studies.

\noindent \textbf{Implementation Details.} For \textit{SimGCD-CLIP + AL-GCD}, we use CLIP ViT-B/16~\cite{radford2021learning} as the backbone. We fine-tune the last layer of both the image and text encoders, and the Fusion-head projector. The Fusion-head projector is an MLP with a 512-dimensional input, a 2048-dimensional hidden layer, and a 256-dimensional output, followed by a GeLU activation~\cite{hendrycks2016gaussian}. We set the learning rate to 0.1. Other hyperparameters, including batch size, temperature \( \tau_s \), weight decay, and the number of augmentations, are set to 256, 0.07, \(1e^{-4}\), and 2. All experiments are conducted on an NVIDIA 3090 GPU.

\begin{table}[t]
\centering
\small
\caption{Comparison on Fine-grained Avg, Classification Avg, and All Datasets Avg with CLIP backbone. The best values are in bold and the second best are underlined. † denotes reproduced results. $\triangle$ denotes results from the CMS appendix \cite{CMS2024}.}
\setlength{\tabcolsep}{1.75pt}
\begin{tabular}{l|ccc|ccc|ccc}
\toprule
\multicolumn{1}{c}{Method} 
& \multicolumn{3}{c}{Fine-grained} 
& \multicolumn{3}{c}{Classification} 
& \multicolumn{3}{c}{All Datasets} \\
\cmidrule(lr){2-4}\cmidrule(lr){5-7}\cmidrule(lr){8-10}
& All & Old & New & All & Old & New & All & Old & New \\
\midrule
\multicolumn{10}{l}{\textit{(a) Clustering with the GT number of categories K given}} \\
\midrule
CPT         & 65.9 & 71.3 & 63.2 & 85.3 & 88.3 & {83.7} & 68.9 & 75.3 & 65.6 \\
GET         & \underline{71.5} & 74.8 & \underline{69.8} & \underline{86.9} & \underline{90.6} & 82.6 & 73.0 & 78.4 & 69.4 \\
CMS $^\triangle$ & 64.7 & 74.5 & 59.9 & 83.1 & 89.5 & 76.3 & 66.1 & 76.5 & 59.6 \\
\rowcolor{gray!15} \quad + AL-GCD & 69.5 & 75.3 & 66.7 & 85.8 & 90.4 & 83.0 & 70.0 & 77.7 & 65.8 \\
SimGCD$^\dagger$ & 64.2 & 71.9 & 60.4 & 85.5 & 89.2 & 81.4 & 68.6 & 76.1 & 64.0 \\
\rowcolor{gray!15} \quad + AL-GCD & 70.5 & \underline{75.8} & 68.0 & \textbf{88.7} & 89.9 & \textbf{88.1} & \underline{73.2} & \underline{78.5} & \underline{70.5} \\
SelEx$^\dagger$ & 66.6 & 65.3 & 67.2 & 84.2& 85.3 & \underline{83.8} & 68.8 & 70.6 & 67.8 \\
\rowcolor{gray!15} \quad + AL-GCD & \textbf{76.6} & \textbf{78.9} & \textbf{75.6} & 86.6 & \textbf{90.9} & 80.5 & \textbf{75.1} & \textbf{79.6} & \textbf{71.5} \\
\rowcolor{gray!30}\textbf{\textit{Avg. Improv.}} 
& \textbf{\textit{+7.1}} & \textbf{\textit{+6.1}} & \textbf{\textit{+7.6}} 
& \textbf{\textit{+2.7}} & \textbf{\textit{+2.4}} & \textbf{\textit{+3.4}} 
& \textbf{\textit{+5.0}} & \textbf{\textit{+4.2}} & \textbf{\textit{+5.5}} \\
\midrule
\multicolumn{10}{l}{\textit{(b) Clustering without the GT number of categories K given}} \\
\midrule
CMS $^\triangle$ & 64.5 & 71.4 & 61.0 & 81.4 & 87.5 & 75.9 & 65.8 & 74.5 & 60.6 \\
\rowcolor{gray!15} \quad + AL-GCD & \textbf{69.6} & \textbf{76.2} & \textbf{66.4} & \textbf{84.5} & \textbf{90.7} & \textbf{80.2} & \textbf{69.8} & \textbf{78.2} & \textbf{65.2} \\
\rowcolor{gray!30}\textbf{\textit{Improvement}}
& \textbf{\textit{+5.1}} & \textbf{\textit{+4.8}} & \textbf{\textit{+5.4}} 
& \textbf{\textit{+3.1}} & \textbf{\textit{+3.2}} & \textbf{\textit{+4.3}} 
& \textbf{\textit{+4.0}} & \textbf{\textit{+3.7}} & \textbf{\textit{+4.6}} \\
\bottomrule
\end{tabular}
\label{tab:comparison}
\end{table}

\newcommand{\cmark}{\checkmark} 
\newcommand{\xmark}{\ding{55}}  
\begin{table*}[t]
\centering
\small 
\setlength{\tabcolsep}{1.7pt} 
\caption{Ablation study results for ATCG with various initial and stacked layer settings across different datasets.}
\begin{tabular}{cc ccc ccc ccc ccc ccc ccc}
\toprule
\multirow{2}{*}{\shortstack{ATCG \\ Initial Layer}} & \multirow{2}{*}{\shortstack{ATCG \\ Stacked Layers}} & \multicolumn{3}{c}{CIFAR100} & \multicolumn{3}{c}{ImageNet100} & \multicolumn{3}{c}{CUB} & \multicolumn{3}{c}{Stanford Cars} & \multicolumn{3}{c}{FGVC Aircraft} & \multicolumn{3}{c}{Herbarium19} \\
\cmidrule(r){3-5} \cmidrule(r){6-8} \cmidrule(r){9-11} \cmidrule(r){12-14} \cmidrule(r){15-17} \cmidrule(r){18-20}
 &  & All & Old & New & All & Old & New & All & Old & New & All & Old & New & All & Old & New & All & Old & New \\
\midrule
\xmark & \xmark & 81.1 & 84.2 & 75.2  & 89.9 & 94.2 & 87.5 & 69.6 & 75.8 & 66.5 & 69.4 & 82.2 & 62.9 & 53.5 & \underline{57.9} & 51.7 & 47.9 & 62.3 & 39.9 \\
\cmark & \xmark & \underline{83.7} & \underline{84.3} & \underline{82.7} &\underline{90.4} & \underline{94.9} & \underline{88.0}  &\underline{73.4} & \underline{76.7} & \underline{71.8} & \underline{74.9} & \underline{86.1} & \underline{69.0} & \underline{56.1} & 56.5 & \underline{55.9} & \underline{48.8} & \underline{62.9} & \underline{41.3} \\
\rowcolor{gray!20} \cmark & \cmark & \textbf{84.7} & \textbf{84.5} & \textbf{84.9} & \textbf{92.6} & \textbf{95.3} & \textbf{91.2} & \textbf{74.7} & \textbf{79.6} & \textbf{72.3} & \textbf{78.3} & \textbf{86.3} & \textbf{74.4} & \textbf{58.6} & \textbf{61.5} & \textbf{57.2} & \textbf{50.3} & \textbf{63.6} & \textbf{43.1} \\

\rowcolor{gray!40} \multicolumn{2}{c}{{\textit{\textbf{Improv. over baseline}}}} & \textbf{+3.6} &\textbf{ +0.3} & \textbf{+9.7} & \textbf{+2.7} & \textbf{+1.1} & \textbf{+3.7} & \textbf{+5.1} & \textbf{+3.8} & \textbf{+5.8} & \textbf{+8.9} & \textbf{+4.1} & \textbf{+11.5} & \textbf{+5.1} & \textbf{+3.6} & \textbf{+5.5} & \textbf{+2.4} & \textbf{+1.3} & \textbf{+3.2}  \\
\bottomrule
\end{tabular}
\label{tab:ablation}
\end{table*}

\subsection{Comparison with State of the Art Methods}
We compare AL-GCD with recent GCD methods \cite{GCD2022,GPC2023,SimGCD2023,SelEx2024,RLCD,CMS2024,CPT,GET}.
Following \cite{CMS2024}, we report two settings: (a) clustering with the ground-truth number of categories and (b) clustering without it.
We plug ATCG into three representative pipelines—SimGCD-CLIP, CMS-CLIP, and SelEx-CLIP.
To ensure direct and fair comparisons, CLIP ViT-B/16 is used for CLIP-based baselines; results for GCD-CLIP and CMS-CLIP are taken from \cite{CMS2024}, while SimGCD-CLIP and SelEx-CLIP are reproduced by us.
We evaluate overall accuracy, known-category accuracy, and novel-category accuracy on six benchmarks.
Per-dataset results are reported in Table~\ref{tab:main}; aggregated averages over fine-grained sets, standard classification sets, and all datasets are in Table~\ref{tab:comparison}.

\noindent\textbf{Classification Datasets.} On CIFAR100 and ImageNet100, attaching ATCG improves overall accuracy for all pipelines.
For instance, SimGCD-CLIP + AL-GCD reaches 84.7\% on CIFAR100 and 92.6\% on ImageNet100; SelEx-CLIP + AL-GCD attains 92.7\% on ImageNet100. Aggregates in Table~\ref{tab:comparison}(a) show average gains on classification datasets of \textbf{+2.7\%} in overall accuracy after attaching ATCG.
When the number of categories is unknown, Table~\ref{tab:main}(b) and Table~\ref{tab:comparison}(b) indicate consistent improvements over CMS-CLIP on the classification aggregate by \textbf{+3.1\%} for all categories, \textbf{+3.2\%} for known categories, and \textbf{+4.3\%} for novel categories.

\noindent\textbf{Fine-Grained Datasets.} On CUB, Stanford Cars, and FGVC-Aircraft, AL-GCD yields substantial gains.
Table~\ref{tab:main}(a) reports strong per-dataset results, such as 84.1\% on CUB and 66.6\% on FGVC-Aircraft with SelEx-CLIP + AL-GCD.
Aggregated fine-grained averages in Table~\ref{tab:comparison}(a) confirm consistent improvements of \textbf{+7.1\%} in overall accuracy, \textbf{+6.1\%} in known-category accuracy, and \textbf{+7.6\%} in novel-category accuracy.
In the setting without the ground-truth number, Table~\ref{tab:comparison}(b) shows further gains over CMS-CLIP of \textbf{+5.1\%} for all categories.

\noindent\textbf{Challenging Dataset: Herbarium19.}
Herbarium19 is a long-tailed, fine-grained benchmark with large per-category sample imbalance.
As reported in \cite{2024zoom}, CLIP achieves only 0.037\% zero-shot accuracy on this dataset, indicating negligible benefit from pretraining memorization.
Table~\ref{tab:main} shows that AL-GCD lifts SimGCD-CLIP to \textbf{50.3\%} overall and \textbf{43.1\%} on novel categories.
Under the unknown-category setting, AL-GCD raises CMS-CLIP to \textbf{41.2\%} overall and \textbf{31.6\%} on novel categories.
These gains demonstrate that analogical visual–textual fusion sharpens novel-category separation on highly imbalanced, fine-grained data.

\noindent\textbf{All Datasets.}
On the aggregated “All Datasets” metrics in Table~\ref{tab:comparison}, attaching ATCG raises the averages by \textbf{+5.0\%} in overall accuracy, \textbf{+4.2\%} in known-category accuracy, and \textbf{+5.5\%} in novel-category accuracy when the category number is given; under the unknown-category setting, the gains are \textbf{+4.0\%}, \textbf{+3.7\%}, and \textbf{+4.6\%}, respectively. These improvements align with the per-dataset results in Table~\ref{tab:main}, indicating that analogical textual concepts consistently sharpen novel-category separation while maintaining strong performance on known categories.

\subsection{Ablation Study}
All ablations are conducted on SimGCD-CLIP + AL-GCD.

\noindent \textbf{The Effectiveness of Each Component.}
To evaluate the contributions of the \textit{Initial Layer} and \textit{Stacked Layers} in the Analogical Textual Concept Generator (ATCG), we conducted an ablation study with three configurations: (1) without ATCG, (2) using only the \textit{Initial Layer}, and (3) incorporating both the \textit{Initial Layer} and \textit{Stacked Layers}. Table~\ref{tab:ablation} summarizes the results across multiple datasets.

\noindent \textit{Initial Layer}: The \textit{Initial Layer}, equipped with the \textit{Text \& Image-Analogical Attention (TIAA)} module, generates meaningful text embeddings by leveraging relationships between known and novel categories. On CIFAR100, adding the \textit{Initial Layer} increases all-category accuracy to 83.7\% and significantly enhances novel category accuracy to 82.7\%, while maintaining stable old-category accuracy. Similar improvements are observed on CUB, where novel category accuracy rises to 71.8\%. These results demonstrate that analogical learning effectively structures novel category representations, improving category separation.

\begin{table}[t]
\centering
\small 
\caption{Ablation study on the number of ATCG layers.}
\vspace{-0.3cm}
\setlength{\tabcolsep}{4.4pt} 
\begin{tabular}{c ccc ccc}
\toprule
\multirow{2}{*}{\shortstack{Num. of \\ ATCG Layers}} & \multicolumn{3}{c}{CUB} & \multicolumn{3}{c}{CIFAR100} \\
\cmidrule(r){2-4} \cmidrule(r){5-7}
 & All & Old & New & All & Old & New \\
\midrule
0 & 69.62 & 75.78 & 66.53 & 81.11 & \underline{84.22} & 75.23 \\
2 & \underline{74.73} & \textbf{79.59} & 72.31 & 84.67 & \textbf{84.54} & 84.94 \\
4 & \textbf{74.80} & \underline{78.48} & \underline{72.87} & \textbf{84.87} & 83.77 & \textbf{86.76 }\\
6 & 74.69 & 78.12 & \textbf{72.97} & \underline{84.76} & 83.67 & \underline{86.63} \\
\bottomrule
\end{tabular}
\label{tab:atcg_layers}
\end{table}

\noindent \textit{Stacked Layers}: Extending ATCG with \textit{Stacked Layers}, which integrate the \textit{TSA} and \textit{TIAA} modules, further refines text embeddings through iterative updates. On ImageNet100, the addition of \textit{Stacked Layers} raises all-category accuracy from 90.4\% to 92.6\%, with novel-category accuracy increasing by \textbf{+3.2\%}. The impact is even more pronounced on fine-grained datasets such as Stanford Cars, where all-category accuracy improves from 74.9\% to 78.3\% and novel category accuracy increases by \textbf{+5.4\%}. These gains highlight the iterative refinement’s role in capturing subtle visual and semantic distinctions, particularly for visually similar categories.

The ablation shows a clear synergy between ATCG’s two types of layers: the \textit{Initial Layer} shapes novel-category embeddings by referencing known categories, and the \textit{Stacked Layers} further refine them for sharper semantic separation. Without ATCG, models struggle with look-alike novel categories; the \textit{Initial Layer} adds structured textual concepts, and the \textit{Stacked Layers} sharpen boundaries. Overall, AL-GCD couples analogical learning with self-supervised textual refinement to improve novel discovery while preserving performance on known categories.

\noindent \textbf{The Influence of the Number of Layers.}  
To evaluate the impact of ATCG depth, we conduct ablations with 0, 2, 4, and 6 layers on CUB and CIFAR100. The results in Table~\ref{tab:atcg_layers} show that iterative refinement improves novel-category discovery and overall performance.
As the number of layers increases, overall and novel accuracies improve consistently up to 4 layers. On CIFAR100, overall rises from 81.1\% (0 layers) to 84.9\% (\textbf{+3.8\%}) with 4 layers, while novel increases from 75.2\% to 86.8\%. Similarly on CUB, overall improves from 69.6\% to 74.8\% (\textbf{+5.2\%}), and novel from 66.5\% to 72.9\%. These results underline the benefit of deeper analogical refinement.
However, beyond two layers, known-category accuracy tends to plateau or slightly decline. This trend reflects a mild trade-off: as new category concepts are refined by analogy to known categories, the optimization focuses more on novel-category, which can slightly affect recognition of known categories.

\begin{figure}[t]
  \centering
  \includegraphics[width=0.85\linewidth]{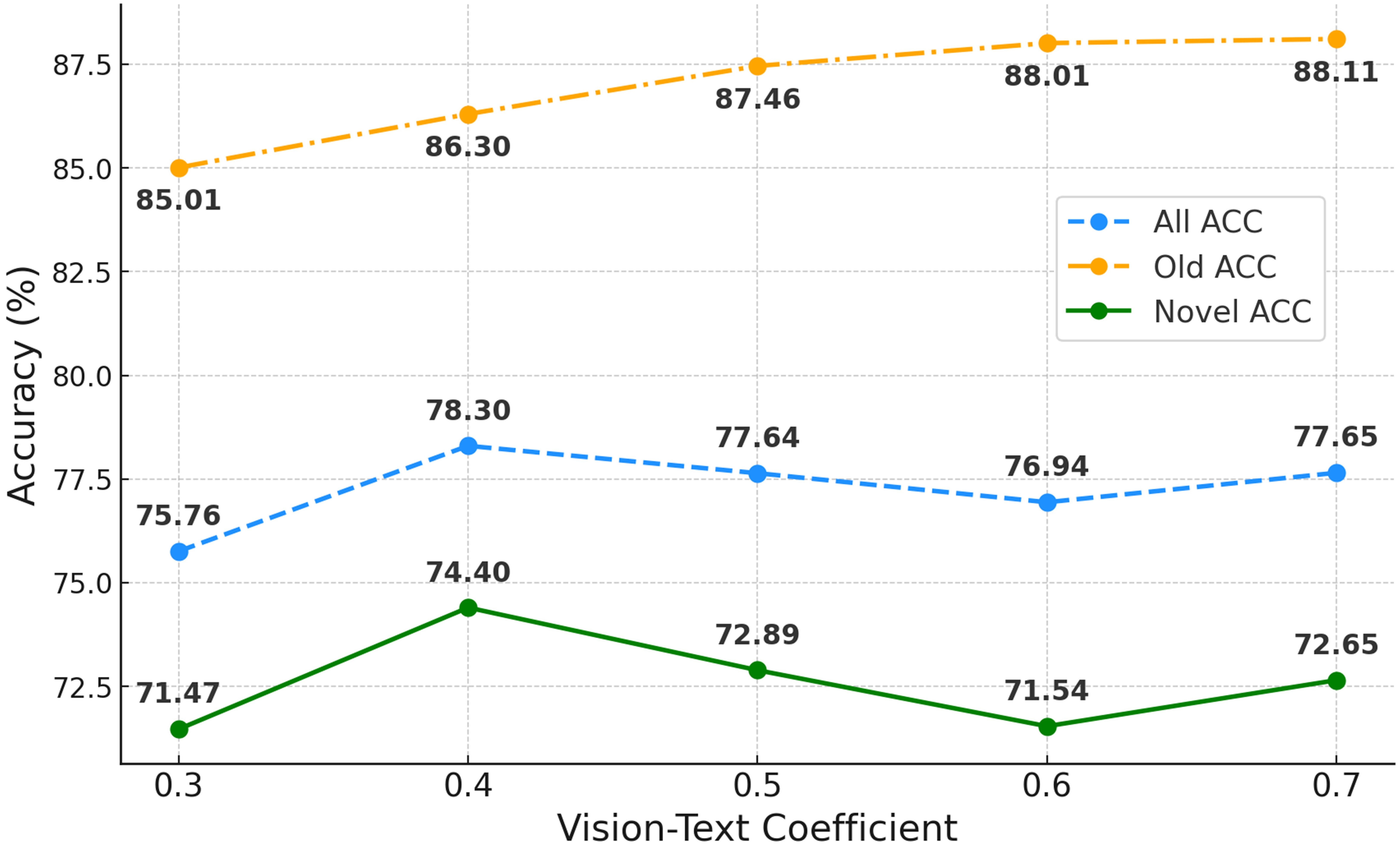}
  \caption{Ablation study of the \( \alpha \) on Stanford Cars.}
  \label{fig:ablation-car}
\end{figure}

\noindent \textbf{The Influence of the Vision-Text Coefficient \( \alpha \).}
The vision-text coefficient $\alpha$ controls the relative contribution of visual and textual information in the fused embedding. As shown in Fig.~\ref{fig:ablation-car}, higher values of \( \alpha \) generally lead to improved old category accuracy but result in a decline in novel category accuracy. On Stanford Cars, when \( \alpha \) increases from 0.4 to 0.7, old category accuracy rises by 1.8\%, while novel category accuracy decreases by 1.7\%. This is because a stronger emphasis on visual features aligns well with the rich representations of old categories but is less effective in discovering novel categories, where textual analogies play a more significant role. Conversely, lower \( \alpha \) prioritizes textual embeddings, enhancing novel category separability. This trade-off highlights the complementary nature of visual and textual information. A balanced value, such as \( \alpha = 0.4 \), achieves strong performance across both old and novel categories by leveraging the complementary strengths of visual and textual information.

\section{Conclusion and Future Work}
In this work, we presented AL-GCD, an analogical learning framework for generalized category discovery built upon the Analogical Textual Concept Generator, which reasons over a visual-textual knowledge base to produce analogical textual embeddings that complement visual features. AL-GCD plugs seamlessly into existing GCD pipelines and consistently improves overall, known-class, and novel-class performance on six benchmarks, with particularly large gains on fine-grained datasets. In future work, we plan to explore more expressive analogical reasoning mechanisms, incorporate richer semantic sources such as large language models, and extend ATCG to continual and open-world discovery scenarios where analogical cues may provide stronger robustness under distribution shifts.

\section{Acknowledgements}
This work is supported by the National Natural Science Foundation of China under Grant No.U21B2048, No.62302382 and No.62576224, the Shenzhen Key Technical Projects under Grant CJGJZD20220517141605013 and the China Postdoctoral Science Foundation No.2024M752584 and No.2025M784372.

{
    \small
    \bibliographystyle{ieeetr} 
    \bibliography{main}
}


\end{document}